\documentclass[letterpaper, 10 pt, conference]{ieeeconf}  

\IEEEoverridecommandlockouts                              
\usepackage{comment}                                                          
\usepackage{mathtools}
\usepackage[final]{pdfpages}
\usepackage{graphicx}
\usepackage{booktabs}
\usepackage{lipsum}
\usepackage{mwe}
\usepackage{graphicx}
\usepackage{subcaption}
\usepackage{multirow}
\usepackage{array}
\usepackage{float}
\usepackage{amsmath}
\usepackage{cite}
\usepackage[utf8]{inputenc}
\usepackage{fourier} 
\usepackage{array}
\usepackage{makecell}
\usepackage{amsmath}
\usepackage{amssymb}

\makeatletter
\let\NAT@parse\undefined
\makeatother

\usepackage{hyperref}

\title{\LARGE \bf
Gradient Aware - Shrinking Domain based Control Design for Reactive Planning Frameworks used in Autonomous Vehicles }

\author{Adarsh Modh$^{1}$, Siddharth Singh$^{1}$, A. V. S. Sai Bhargav Kumar$^{1}$, Sriram N. N.$^{1}$, K. Madhava Krishna$^{1}$
\thanks{$^{1}$ The authors are with Robotics Research Center, IIIT-Hyderabad India. Equal Contribution from the first two authors.
{\tt\small adarsh.modh@gmail.com}
{\tt\small singh.sid930@gmail.com}
{\tt\small mkrishna@iiit.ac.in}        
}%
}

\begin{document}

\maketitle
\thispagestyle{empty}
\pagestyle{empty}

\begin{abstract}
In this paper, we present a novel control law for longitudinal speed control
of autonomous vehicles. The key contributions of the proposed work include the design of a control law that reactively integrates the longitudinal surface gradient of road into its operation. In contrast to the existing works, we found that integrating the path gradient into the control framework improves the speed tracking efficacy. Since the control law is implemented over a shrinking domain scheme, it minimizes the integrated error by recomputing the control inputs at every discretized step and consequently provides less reaction time. This makes our control law suitable for motion planning frameworks that are operating at high frequencies.
Furthermore, our work is implemented using a generalized vehicle model and can be easily extended to other classes of vehicles. The performance of gradient aware - shrinking domain based controller is implemented and tested on a stock electric vehicle on which a number of sensors are mounted.
Results from the tests show the robustness of our control law for speed tracking on a terrain with varying gradient while also considering stringent time constraints imposed by the planning framework.

\end{abstract}

\section{INTRODUCTION} \label{section1}

The development of Autonomous Vehicles in the recent times has pushed research in the multi-disciplinary domain of robotics by integrating concepts of control theory, motion planning framework, vision and perception. In the proposed work, the efforts are made towards the development of a control framework for robust speed tracking of autonomous vehicles which is in coherence with the planning and perception/vision framework. For real time tracking of planning framework and perception inputs in real-world scenarios, it becomes imperative that there are no bottlenecks in the pipeline which can slow down the system and may produce integrated errors over time. It is seen that the conventional low-level control frameworks tend to slow down the planning computations due to their interactions with the system and inability to capture the exact non-linear dynamics of the vehicle model.

In the proposed approach however, the focus is on the improvement of the control framework for speed tracking which allows the existing pipeline of the autonomous vehicle to execute operations at a high rate without losing out on the robustness. The framework has been implemented and tested on an Electric Vehicle, Mahindra e2o, the requisite vehicle parameters of which are known. The control framework is able to handle terrains with undulating slopes and surfaces in real time by incorporating sensor data. Gradient control is made possible by integrating real time data from the IMU sensor. Friction of the surface is modeled through widely used PAC2002 model \cite{kuiper2007pac2002} by calculating longitudinal slip in real time through the IMU sensor and quadrature incremental optical wheel encoders. 

Comparison between the proposed framework and tuned PID framework which is conventionally used in autonomous vehicles for speed tracking is presented. During the tests on PID it is found that the tuning parameters modeled over a flat surface does not show the same performance over gradient and other frictional surfaces and vice versa. Moreover, tuned PID over a particular gradient angle seems to perform poorly over other different gradients of varying slopes. 

\begin{figure}[h]
\centering
\includegraphics[width=0.45\textwidth]{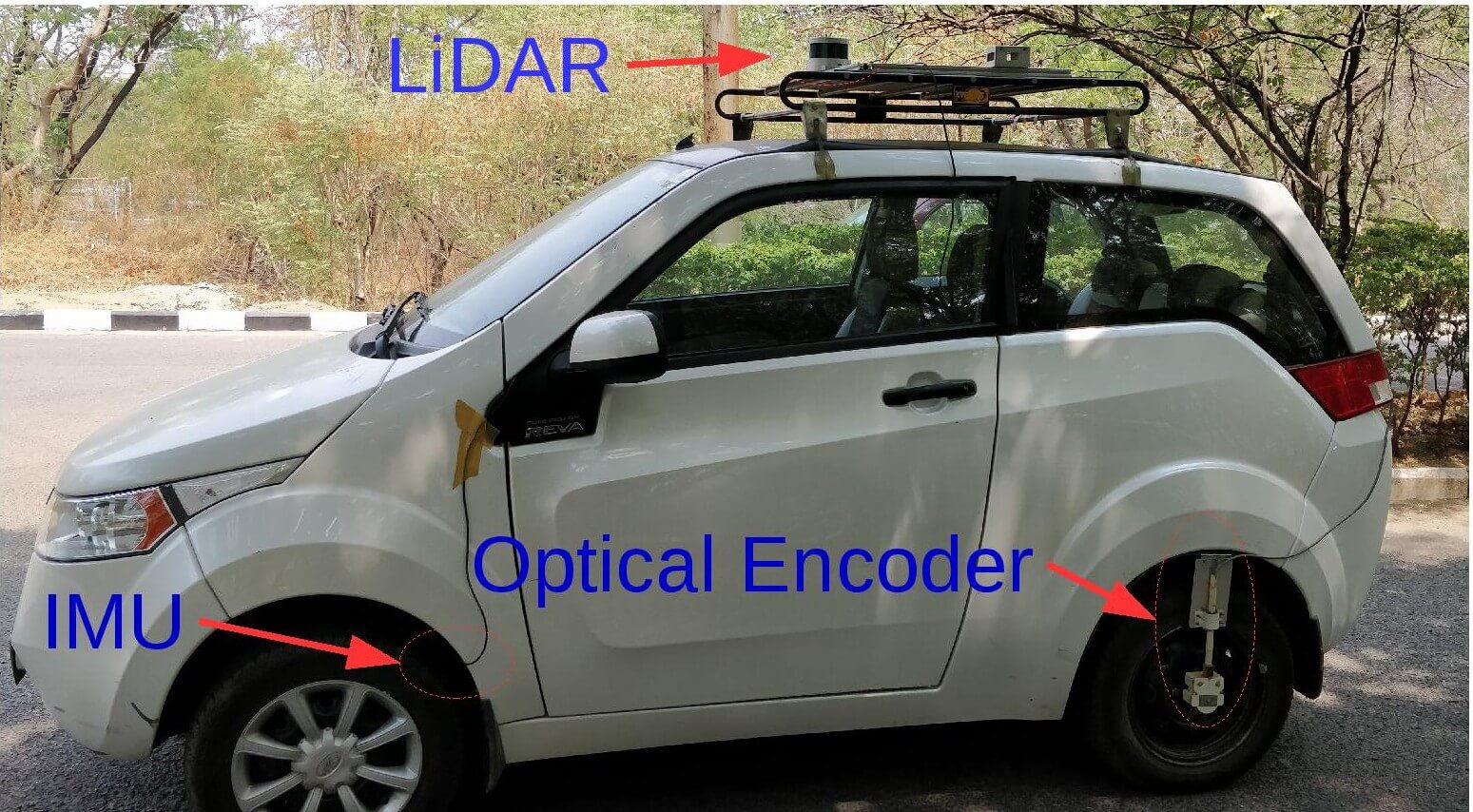}
\caption{Autonomous Vehicle Used for Testing}
\label{setup}
\end{figure}

The rest of the paper is organized as follows. A overview of the work in this field is presented in Section\ref{section2}. Section\ref{section3} describes the entire control framework of our proposed approach. A detailed evaluation of our framework is presented in Section\ref{section4}. Finally, the conclusion and the future work are presented in Section\ref{section5}.

\section{Related Work} \label{section2}
	In this section, we discuss the existing methods that have attempted to develop control laws for speed tracking. The initial work includes the development of control laws for vehicle following at both low speed and high speed scenarios \cite{801226}. A non-linear vehicle system is modeled to obtain the dynamic solution with safety and comfort constraints in \cite{4105943}. In \cite{choi2002hybrid},\cite{antsaklis2000brief},\cite{decarlo2000perspectives} they even explored the hybrid system framework, where the vehicle system is divided into local subsystems.   

	A model-based task automation system design is proposed in \cite{1227578} by taking perceptual states and triggering events. In \cite{shakouri2011adaptive} the vehicle model is linearized for gain scheduling design of a linear quadratic controller for both throttle and brake actuations. The vehicle parameters used for designing and tuning for the Adaptive cruise control(ACC) integrated with collision avoidance strategy is proposed  in \cite{7795572}. But the above mentioned approaches work only on accurate vehicle models and their performance degrades when the parameters deviate from the modeled values.
    
	Model free techniques like intelligent PID controller and a fuzzy controller are implemented in \cite{5625228}. A fuzzy iterative learning based control is presented in \cite{7531726} which can work on MIMO systems with variable initial errors.  But all the approaches presented above does not consider the road gradient which would effect the controllers performance considerably.   
    
    Significant work is done in  \cite{kim2016time} with respect to speed tracking for vehicles and they too perform their analysis over gradients. However, their system requires initial prior estimation of certain coefficients, tuning parameters and adaptation gains needs to be determined using the knowledge of the power-train system and the braking system. For their purpose they employ a linearized longitudinal vehicle model coupled with a transient parameter adaptive control algorithm. Another work, \cite{chen2017adaptive} makes use of a single neuron neural network coupled with a PID controller based on a quadratic criterion. The method performs identification for the model for simulation experiments to derive the initial parameters of the controller and further test the performance on a miniature autonomous vehicle. 
    
Furthermore, for modeling the road condition, in \cite{chen2011adaptive} a methodology for estimating the tire-road friction coefficient in real time is proposed using the LuGre-type dynamic tire model. They perform their tests through simulations on a high fidelity Car-Sim full-vehicle model. On the front of collision avoidance in autonomous vehicles, \cite{moon2009design} have proposed a Adaptive control system which can also perform collision avoidance. They have been able to implement the system on a real vehicle and test it in safe traffic and severe-braking situations. They also provide results for high speed driving and low speed urban driving with stop-and-go situations. 

Our approach is novel in its unique integration of road gradient in the control formulation and solving it in a shrinking domain framework which reduces the tracking error even when the vehicle model is uncertain and also reduces the computation time.
    
\section{ Control Framework Setup } \label{section3}

\subsection{Vehicle Model-Longitudinal Dynamics}

The performance of the speed tracking framework largely depends on the modeling of the vehicle and its dynamic parameters which are non-linear in nature. In the presented control framework, both the planning level constraints as well as the current kino-dynamic constraints are considered for the speed control of the vehicle. It is important to capture the vehicle dynamics without making the computation too intensive or making the response time too high for it to react. A simplified longitudinal generic vehicle model \cite{rajamani2011vehicle}  which is coupled with the motor characteristics of the electric vehicle described below and the shrinking domain control described in the section after that. Coupling the motor characteristics with the vehicle dynamics helps in capturing the non-linear dynamics of the system model for varying load-RPM conditions. Implementing this system over a shrinking domain framework helps in error reduction.

\begin{figure}[h]
\centering
\includegraphics[width=0.45\textwidth]{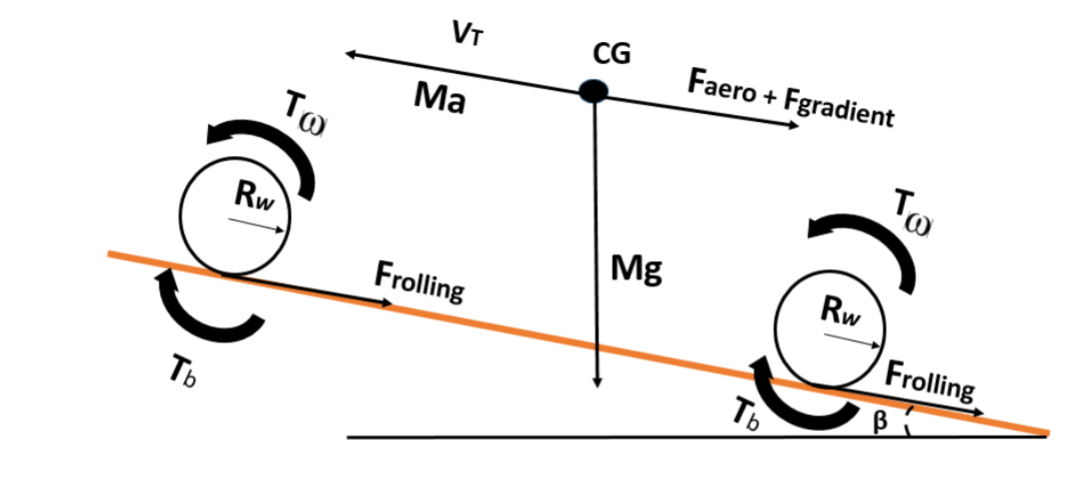}
\caption{Longitudinal Vehicle Model}
\label{Long_VM}
\end{figure}

The vehicle model coupled with the motor characteristics is used for developing the control laws and system equations. In longitudinal frame of the vehicle the forces acting are represented in Fig.\ref{Long_VM}. The driving force being generated from the motor is essentially the torque being delivered to the wheels and consequently the force at the tire-ground contact patch to move the vehicle. The hindering forces are generated from the ground conditions, the gradient and the air viscosity \cite{gillespie1992fundamentals}.\\ 

The acceleration of the vehicle can be computed from Newton's second law of motion- 

\begin{equation}\label{eq:Longitudinal vehicle dynamics}
\begin{gathered}
 \frac{dv}{dt} = \frac{1}{m}(F(t)_{driving} - F(t)_{hindrance})
  \end{gathered}
 \end{equation}

where, \\
${m}$ is the overall mass of the vehicle \\
${v}$ is the longitudinal velocity of the vehicle \\ ${F(t)_{driving}}$ is the time varying driving force being delivered to the vehicle's tires \\ ${F(t)_{hindrance}}$ are the time varying hindering forces arising from external disturbances \\

The driving forces and hindering forces can be represented as follows- 
 \begin{equation}\label{eq:Rolling Resistance}
\begin{gathered}
 F_{rolling}(t)=\mu_{rolling}  M  g  cos\theta
  \end{gathered}
 \end{equation}
 \begin{equation}\label{eq:Gradient Resistance}
\begin{gathered}
 F_{gradient}(t)= M  g  sin\theta
   \end{gathered}
 \end{equation}
  \begin{equation}\label{eq:Aerodynamic Resistance}
\begin{gathered}
 F_{aero}(t)= \frac{1}{2}  \rho  A  v^2  C_d
  \end{gathered}
 \end{equation}
   \begin{equation}\label{eq:Longitudinal dynamic resistance}
\begin{gathered}
 \boldsymbol{F_{hindrance}(t)= F_{rolling}(t) + F_{gradient}(t) +  F_{aero}(t)}
  \end{gathered}
 \end{equation}

  \begin{equation}\label{eq:Longitudinal driving force}
\begin{gathered}
\boldsymbol{F_{driving}(t)= \frac{T_w(t)}{R_{eff}}- \frac{T_{b}(t)}{R_{brake}}}
  \end{gathered}
 \end{equation}

  \begin{equation}\label{eq:Motor Torque}
\begin{gathered}
 T_{m}(t)= T_w(t) * \frac{1}{r_g} * \frac{p_{loss}}{100}
  \end{gathered}
 \end{equation}

where,\\
${T_w(t)}$ is the torque obtained at the wheels\\ ${T_b(t)}$ is the Brake torque\\
${T_m(t)}$ is the Motor (driving) torque\\ 
${r_g}$ is the gear ratio for the included gearbox in the vehicle \\
${p_{loss}}$ is the percent loss in the torque due to frictional and heat losses arising in the gearbox and differential of the vehicle \\
$R_{eff}$ is the effective tire radius \\
$R_{brake}$ is the effective brake radius \\

For mapping the various hindering forces we analyze the parameters which affect them. For a fast and robust response the hindering forces are calculated in real time using these parameters. The separate hindering forces are as follows:

1. Rolling friction through Pacejka model \cite{kuiper2007pac2002} and vehicle longitudinal slip

2. Gradient using IMU

3. Aerodynamic force from the current velocity

However, at low speeds we can neglect the Aerodynamic forces (${F_{aero}}$) creating the hindering forces of drag. 

From the given system equations we derive the torque required for achieving the velocity in the given time frame as required by the planning framework. Based on the torque requirement and the target velocity requirements, the motor characteristics formulation computes the needed voltage/pedal angle to be delivered to the motor in a discrete time interval over the shrinking domain. 

For mapping the brake torque, the vehicle's tire was allowed to rotate at different free wheel angular velocities and different brake pedal angles relating to various brake percentages were applied. Thus, the resulting torque characteristics was mapped relating to initial ${\omega}$ and the applied brake percentage.

\subsection{3-phase Induction Motor Sub-system}

As mentioned earlier in equation\eqref{eq:Motor Torque}, the torque obtained at the wheels of the vehicle is a factor of the torque generated by the induction motor. A 3-phase squirrel cage induction motor is used for driving the electric vehicle. In order to drive this motor, a 3-phase power supply is required which is provided by an inverter circuit that converts the 48V DC supply from the battery backup. This is a Voltage Source Inverter (VSI) \cite{mohan1995power} \cite{dubey2002fundamentals} circuit consisting of 6 IGBTs (Insulated Gate Bipolar Transistor) and the firing angles for each of those are provided from a separate Control Circuit. This Control Circuit generates SPWM (Sinusoidal Pulse Width Modulating) signals which are fed as gate pulses to the IGBTs. The accelerator pedal has a potentiometer embedded inside it, which acts as a voltage divider. A voltage in the range of 0-5V from the accelerator pedal acts as an input to the control circuit of the VSI to vary the speed of the induction motor. For performing the speed control of the 3-phase induction motor, $V/f$ control method (ratio of line voltage to supply frequency is always constant) is implemented \cite{dubey2002fundamentals}. Thus, it is the task of the Control Circuit to generate the SPWM signals in such a way that both the line voltage as well as the frequency supplied to the induction motor are varied along with maintaining the $V/f$ ratio. Amongst many other methods for speed control of 3-phase induction motor \cite{fitzgerald2003electric}, the major advantage of using the $V/f$ method is that it can safely vary the speed over the entire range of slip, without the motor going into saturation (unlike other methods which are applicable only in limited ranges of slip). This is very crucial in the context of its application to an electric vehicle as the speed of the vehicle has to vary continuously. Also varying the supply frequency for controlling the speed of the motor is convenient due to the Voltage Source Inverter.    

\begin{figure}[h]
\centering
\includegraphics[width=0.45\textwidth]{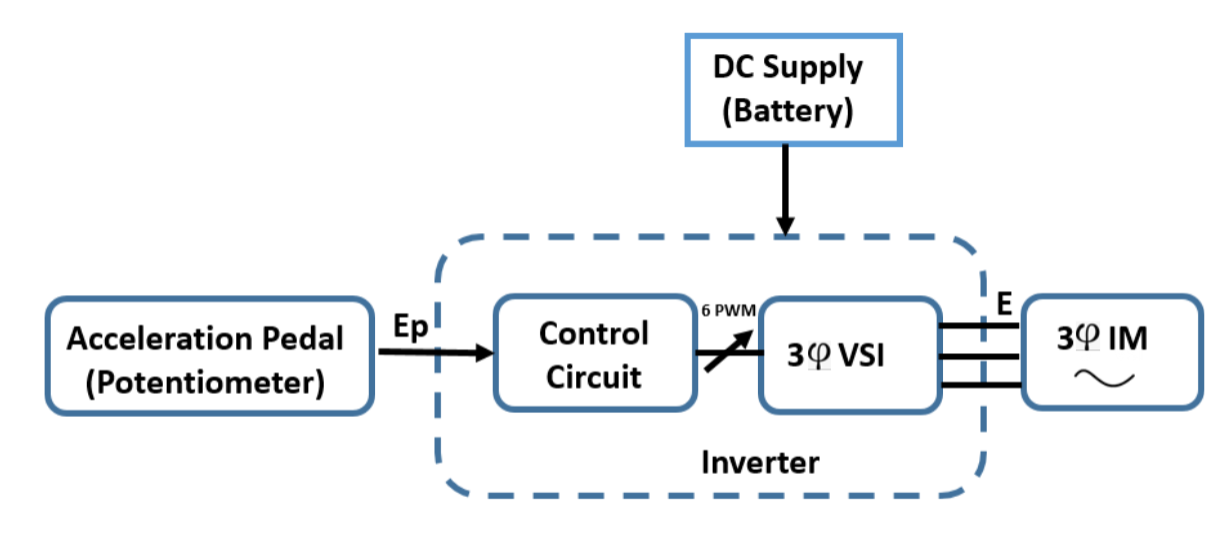}
\caption{Block Diagram of $V/f$ control of VSI fed 3-phase Induction Motor}
\label{IM_block}
\end{figure}

The speed in RPM of the 3-phase induction motor\cite{fitzgerald2003electric} is given by  
\begin{equation}\label{eq:speed_im}
\begin{gathered}
N = \frac{120f}{P}
 \end{gathered}
 \end{equation}
where, \\
$N$ is the speed in RPM of the motor at a given instant\\
$f$ is the frequency of the voltage supplied to the motor\\
$P$ is the number of poles of the motor \\

The Torque-slip characteristics of a 3-phase induction motor \cite{fitzgerald2003electric} is given by

\begin{equation}\label{eq:torque_slip}
\begin{gathered}
T_m =  s E^2\frac{R}{R^2+(sX)^2}\frac{3}{2 \pi N_s}
 \end{gathered}
 \end{equation}

where, 
$T_m$ is the torque produced by the motor
$s$ is the slip which is defined as 
\begin{equation}\label{eq:slip}
\begin{gathered}
s = 1-\frac{N}{N_s}
 \end{gathered}
 \end{equation}
$N_s$ is the synchronous speed (maximum speed) of the 3-phase induction motor\\
$E$ is the voltage supplied to the motor\\
$R$, $X$ are the resistance and the impedance in the rotor circuit of the induction motor\\

$N$ is a function of angular speed $\omega$ at the shaft of the motor which can be expressed as 
\begin{equation}\label{eq:rpm_omega}
\begin{gathered}
N = \omega \frac{60}{2\pi}
 \end{gathered}
 \end{equation}

From equation\eqref{eq:rpm_omega}, equation\eqref{eq:slip} can be written as 
\begin{equation}\label{eq:slip_redefined}
\begin{gathered}
s  = 1 -  k_{2} \omega 
 \end{gathered}
 \end{equation}

where, $k_2 = \frac{60}{2 \pi N_s}$\\

Under operating conditions, $R >> sX$, which implies, 
\begin{equation}\label{eq:R2_redefined}
\begin{gathered}
R^2 + (sX)^2 \approx R^2
~~~\therefore \frac{R}{R^2+(sX)^2} \approx \frac{1}{R}
\end{gathered}
\end{equation}

Also from the design characteristics of the inverter, $E$ is directly proportional to the output voltage $E_p$ of the potentiometer inside the accelerator pedal, which is also the input command to inverter.

\begin{equation}\label{eq:inverterpot}
\begin{gathered}
E = k_e E_p
 \end{gathered}
 \end{equation}
 
where, $k_e$ is a constant.\\
From equations\eqref{eq:torque_slip}\eqref{eq:slip_redefined}\eqref{eq:R2_redefined}\eqref{eq:inverterpot},

\begin{equation}\label{eq:torque_slip_redefined}
\begin{gathered}
T = (1 - k_2 \omega) (k_e E_p)^2 \frac{1}{R}\frac{3}{2 \pi N_s}
 \end{gathered}
 \end{equation}

Equation\eqref{eq:torque_slip_redefined} can be written as
\begin{equation}\label{eq:torque_final}
\begin{gathered}
T = k_1 {E_p}^2 (1 - k_2 \omega)
 \end{gathered}
 \end{equation}

where, $k_1 = \frac{3 {k_e}^2}{2 \pi N_s R}$\\

From the data given by the manufacturer of the induction motor and the inverter

\textit{$k_1 = 0.06692 , k_2 = 0.00126$}

Consequently, $E_p$ is directly proportional to the angle of the acceleration pedal.
Hence, for the given vehicle, the torque obtained at the wheels of the vehicle, which in itself is directly proportional to the torque provided by the motor, is a function of the accelerator pedal angle and the angular speed of the wheel which in turn is directly proportional or equivalent to $\omega$.

\subsection{Shrinking Domain Control}

In the proposed approach, initially, an MPC framework \cite{del2010automotive} was integrated which provided a receding horizon control. However, MPC with it predictive nature proves to be computational intensive and thus a transient shrinking domain control was suggested which is similar to receding horizon control but does not perform the future prediction of the input and state variables. The results suggest that we are still able to reduce errors by a considerable extent and also save on the computation time which directly helps the planning framework in maintaining its re-planning capabilities to a high frequency and least computation. Conventionally, the controller acts as a bottleneck for the planning framework working at lower frequencies and thus the planning framework has to perform repeated tasks of re-planning and computation taking up valuable reaction time in the system. The proposed approach however reduces the bottleneck considerably and provides the planning framework algorithms to work at a much faster rate.

\begin{equation}\label{eq:horizon time steps}
\begin{gathered}
t = \Big[T\bigg(1-\frac{i}{n}\bigg)\Big]
 \end{gathered}
 \end{equation}
 ${Rewriting~\eqref{eq:Longitudinal vehicle dynamics}~in~descretized~state~and~ including~}$
 ${values~of~F_{hinderance}~and~F_{driving}}$
 \begin{equation}\label{eq:horizon time steps discretized}
\begin{gathered}
 \frac{V_T-U}{t} = \frac{\frac{T_w}{R_w}-\frac{T_b}{R_b}-F_{gradient}-F_{rolling}}{M}
 \end{gathered}
 \end{equation}
 \begin{equation}
\begin{gathered}
V_T-U = \Big[T\bigg(1-\frac{i}{n}\bigg)\Big]\Bigg[\frac{\frac{T_w}{R_w}-\frac{T_b}{R_b}-F_{gradient}-F_{rolling}}{M}\Bigg]
 \end{gathered}
 \end{equation}
  \begin{equation}
 \begin{gathered}
\frac{T}{R_w} - \frac{T_b}{R_b} = M\Bigg(\frac{V_T-U}{T(1-\frac{i}{n})}\Bigg) + F_{gradient} + F_{rolling} 
 \end{gathered}
 \end{equation}
 
 A function \eqref{eq:discretization function} dependent on discretization and current velocity thus needs to be solved in real time to derive the torques of the acceleration and brake separately. The function will be solved over a discretized state of ${n}$ discretizations. 
 
\begin{equation} \label{eq:discretization function}
 \begin{gathered}
f(U,i)= {\Bigg[M\Bigg(\frac{V_T-U}{T(1-\frac{i}{n})}\Bigg) + F_{gradient} + F_{rolling}\Bigg]}_{i=0}^{i=n}
 \end{gathered}
 \end{equation}
 
 It is also observed and intuitive to understand that the braking and accelerating scenarios are exclusive and can be targeted separately.
 
 On the basis of the hindering force and the required inertial force we can derive a system which switches between the braking and accelerating mode as per the vehicle's parameter and the current and target velocity.

\begin{equation}\label{eq:risk_pdf_collcone}
\frac{V_T-U}{t} =
\begin{cases}
&  \frac{\frac{T_w}{R_w}-F_{gradient}-F_{rolling}}{M}\\ 
 & ~~~~~~~~~if ~~   m(\frac{V_T-U}{t})+F_{gradient}+F_{rolling} > 0 \\
&  \frac{-\frac{T_b}{R_b}-F_{gradient}-F_{rolling}}{M}\\ 
 & ~~~~~~~~~if ~~   m(\frac{V_T-U}{t})+F_{gradient}+F_{rolling} < 0 \\
\end{cases}
\end{equation}

  \begin{equation}
 \begin{gathered}
T_w=0~,~T_b =0\\~~~~~~~if~m(\frac{V_T-U}{t})+F_{gradient}+F_{rolling} = 0 \\
 \end{gathered}
 \end{equation}

\subsection{Combined System and Algorithm}

The previous sections explain the system model of the vehicle and the three phase induction motor which are to be integrated together over the transient decreasing domain control. An algorithm of process flow can be derived by combining the derived models of the system. 

\begin{figure}[H]
\includegraphics[width=0.45\textwidth]{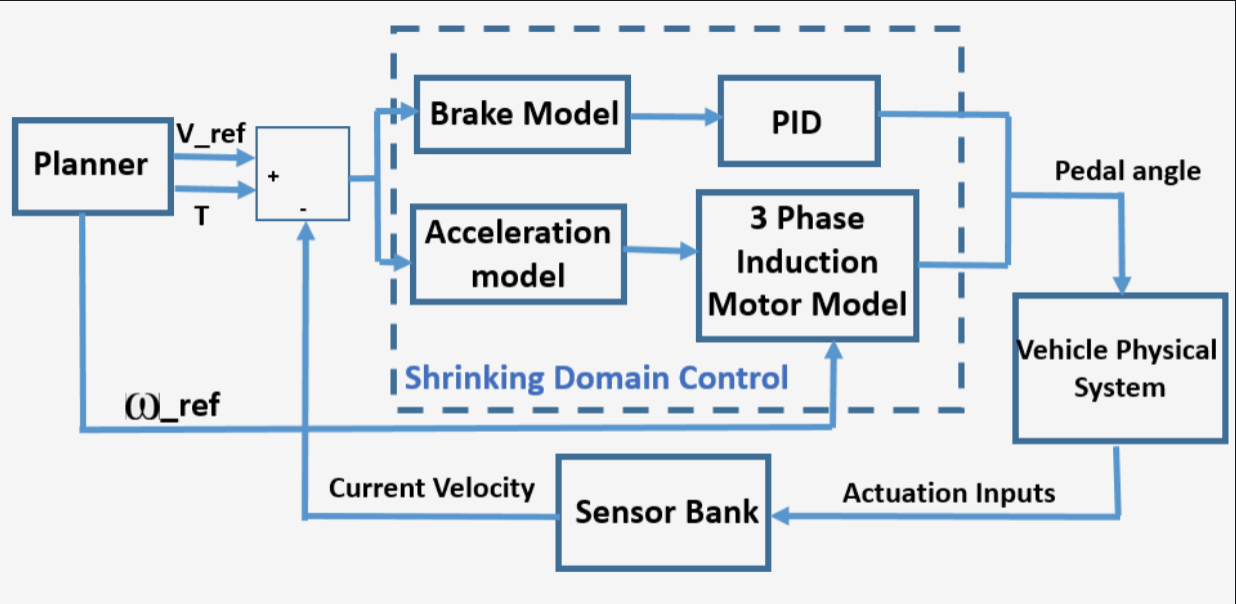}
\caption{Complete Pipeline of the Proposed Work}
\label{total_block}
\end{figure}

In the given Algorithm:

$V_T$: Target velocity\\
$a_{ul},a_{ll}$: Acceleration bounds\\
$T$: Time step \\
$n$: Discretization steps\\
$r_g$: Gearbox ratio\\
$P_{loss}$: Power loss in vehicle transmission\\
$K_1,K_2$: Motor coefficients

\noindent\rule{8cm}{0.9pt}

$\textbf{Algorithm}$

\noindent\rule{8cm}{0.9pt}

1.${\textbf{Initialization}: V_T,(a_{ul},a_{ll}),T,n,i=0,r_g,}$

~~~~~~~~~~~~~~~~~~~~ ${p_{loss}, K_1, K_2}$

\vspace{0.25cm}

2.${\textbf{Initialize~time~update~variable} :~ t=T\Big[1-\frac{i}{n}\Big]}$

\vspace{0.25cm}

3. ${\textbf{while}~(T-t)>0}$ \textbf{do}

\vspace{0.25cm}

4. ~~~~~~~${Update~State~Variables:\theta,U_{current},a,\mu_r,s}$

\vspace{0.25cm}

5. ~~~~~~~${Update~Force~Variables: F_{gradient}, F_{rolling}}$

\vspace{0.25cm}

6. ~~~~~~~${F_{hinderance}= F_{gradient}+F_{rolling}}$

\vspace{0.25cm}

7. ~~~~~~~${a_{ll}<|\frac{U-V_T}{t}|<a_{ul}}$

\vspace{0.25cm}

8. ~~~~~~~${\textbf{if}~~m\frac{V_T-U}{t}+F_{hinderance}>0}$

\vspace{0.25cm}

9. ~~~~~~~~~~~~~~${T_w=R_w(m(\frac{V_T-U}{t})+F_{hinderance})}$

\vspace{0.25cm}

10.~~~~~~~~~~~~~~${T_m=\frac{T_w}{r_g*p_{loss}}}$

\vspace{0.25cm}

11.~~~~~~~~~~~~~~${accelerator~pedal~angle=\frac{T_m}{k_1(1-\omega*k_2)}}$

\vspace{0.25cm}

12.~~~~~~~ ${\textbf{else if}~~m\frac{V_T-U}{t}+F_{hinderance}<0}$

\vspace{0.25cm}

13.~~~~~~~~~~~~~~${T_b=-R_b(m(\frac{V_T-U}{t})+F_{hinderance})}$

\vspace{0.25cm}

14.~~~~~~~~~~~~~~${Brake~pedal~angle=k_{lookup}(T_b-T_{map})}$

\vspace{0.25cm}

15.~~~~~~~ ${\textbf{else if}~~m\frac{V_t-u}{t}=F_{hinderance}}$

\vspace{0.25cm}

16.~~~~~~~~~~~~~~ ${T_w=0}$

\vspace{0.25cm}

17.~~~~~~~~~~~~~~ ${T_b=0}$

\vspace{0.25cm}

18.~~~~~~~ ${\textbf{end~if}}$

\vspace{0.25cm}

19. ${\textbf{end~while}}$

\noindent\rule{8cm}{0.9pt}

\section{Results} \label{section4}
We evaluated the performance of the proposed by integrating our controller for an autonomous vehicle and tested it in different scenarios. The results obtained were compared with a standard PID controller. These testing scenarios include varying the velocity profiles at different gradient conditions as encountered in urban driving scenarios. The robustness of the controller is depicted by the accuracy of the vehicle in tracking the provided velocity profile from the planner. Scenarios like set point tracking, continuous velocity tracking and Stop and go scenarios were used as benchmarks to evaluate our framework on the autonomous car for both flat and gradient surfaces as shown in Section\ref{plots}. The velocity profiles for these scenarios are generated from a higher level planning framework and are transferred to the lower level controller. The sensors integrated with the system helps in providing the information required by the low level controller to perform the computations. This information includes the current velocity computation from the wheel encoders and the gradient of the surface sensed from the IMU. The velocity response with reference to time is recorded from the vehicle by using the same sensors for further analysis. Furthermore, an empirical analysis to provide a quantitative comparison of parameters like the rise time, the root mean square error and the steady state error of the response of the proposed controllers and the standard PID is also presented.

The PID controller being compared is tuned through the performance analysis on a flat surface. It is seen that the gains of the PID controller need to change with the gradient of the surface, so no particular gain will be useful for all kind of terrains. This is where the proposed approach excels as it does not require  prior tuning of any parameters for different terrains and performs the necessary computations to derive control inputs in real time.

The autonomous car uses an indigenous Time Scaled Collision Cone(TSCC) planning framework\cite{2017arXiv171204965B}  developed by using the concepts of time scaling and velocity obstacle. The scale optimization layer of the planning framework solves the time scaled collision cone constraint reactively to obtain the collision free velocities. This optimization problem is constrained by the velocity and acceleration bounds of our autonomous vehicle, thus ensuring the velocity profiles that the planning framework generates are always under the achievable actuation limits of our autonomous vehicle.

All the tests have been performed on the Mahindra e2o electric vehicle. Table\ref{tab:vehicle parameters} shows the required vehicle parameters that were used during testing and computation.
Data is collected under varying dynamic conditions on the vehicle through the following 
sensors: \\
1. LiDAR Velodyne VLP-16\\
2. Incremental Quadrature Optical Encoders HEDS 5645\\
3. Xsense IMU MTi-30

\vspace{0.1cm}
\captionof{table}{Vehicle Parameters} \label{tab:vehicle parameters}
\begin{center}
\begin{tabular}{l*{6}{c}r}
\hline
Parameter              & Value \\
\hline
Mass($M$) &1250Kg   \\
Wheel Radius($R_w$) &0.27m  \\
Brake Radius &0.14m  \\
Rolling Resistance($\mu_r$) &0.025-0.03(for dry asphalt) \\
Gear Ratio($r_g$)  & 10.23 \\
Power Loss($P_{loss}$) & 0.85 \\
\hline
\end{tabular}
\end{center}

\vspace{0.1cm}

\begin{figure}[h]
\includegraphics[width=0.48\textwidth]{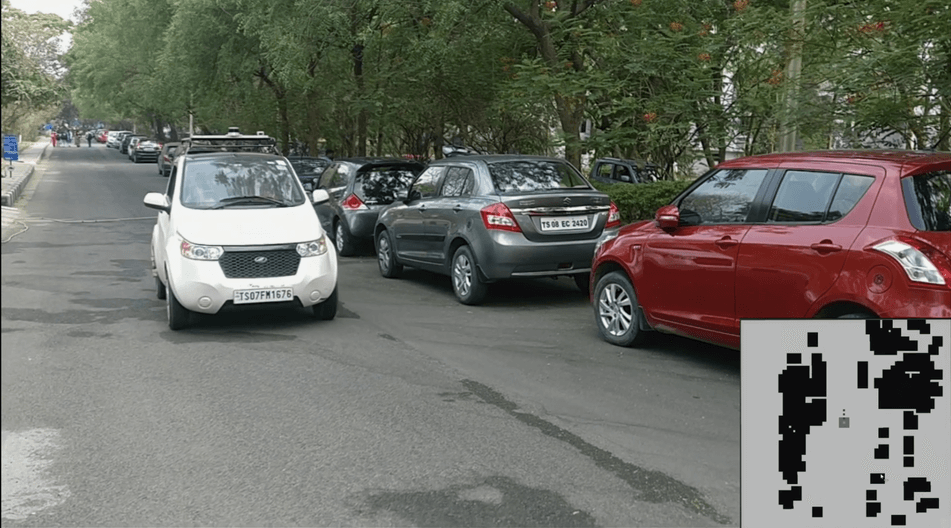}
\caption{Snapshot of the implementation of the proposed control model on Mahindra e2o, overlaid with the occupancy grid map of the surroundings generated from the LiDAR point cloud information which is used by our TSCC planning framework for generating the reference trajectory and its velocity profile}
\label{total_block}
\end{figure}

\subsection{Experimental Results on the Vehicle} \label{plots}
\textit{For all the results shown in Fig.\ref{set_point},\ref{continuous_rise},\ref{stopngo}, the blue line is the reference velocity profile provided by the planning framework, the red line is the velocity profile tracked by the PID controller and green line is the velocity profile tracked by the proposed controller.} 

\subsubsection{Set-point Velocity Tracking} 
\begin{figure}[H] 
    \begin{subfigure}[b]{0.48\linewidth}
        \includegraphics[width=\textwidth,height = 3.5cm]{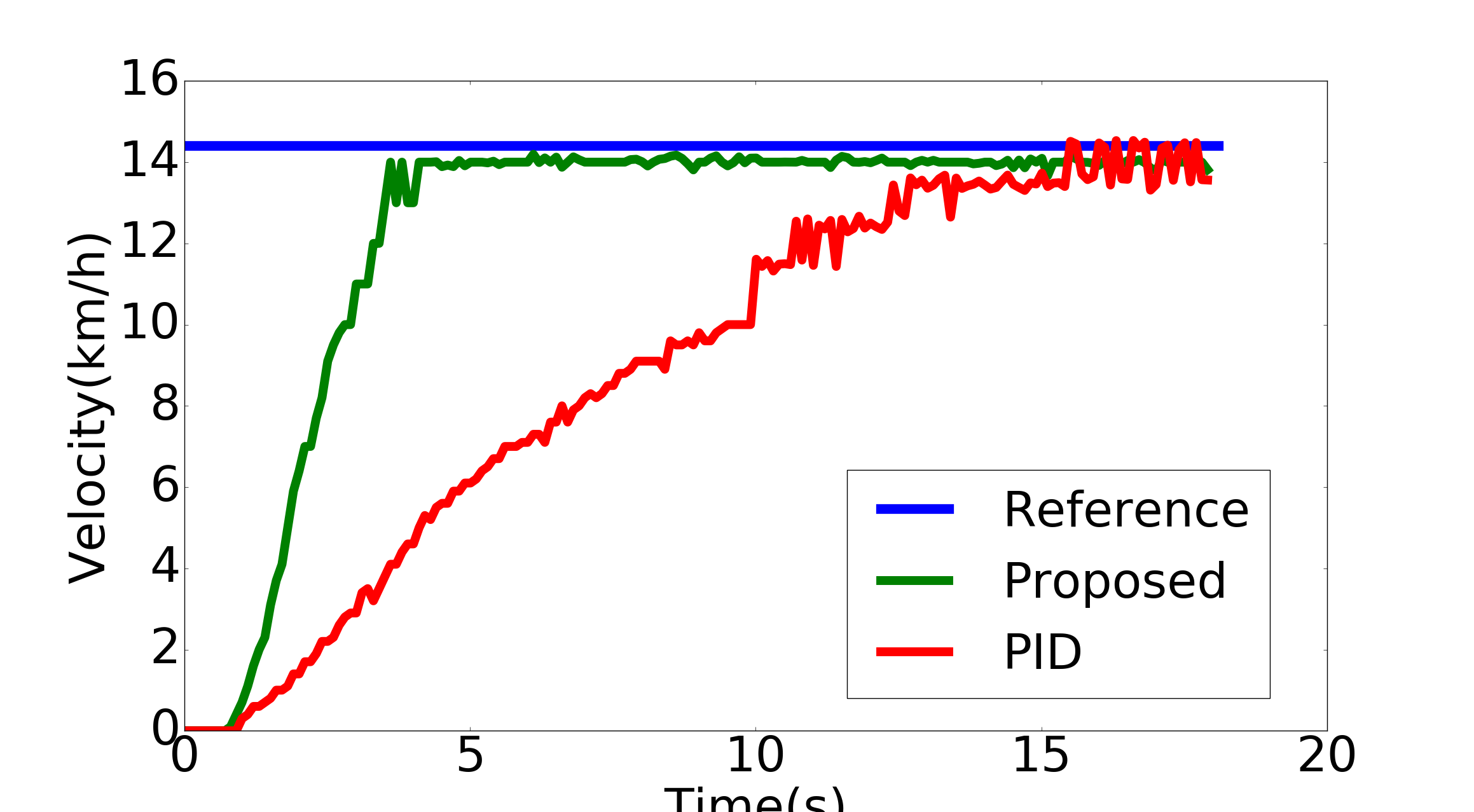}
        \caption{Performance over\\Flat-surface}
        \label{sp_flat}
    \end{subfigure}
    \begin{subfigure}[b]{0.48\linewidth}
        \includegraphics[width=\textwidth,height = 3.5cm]{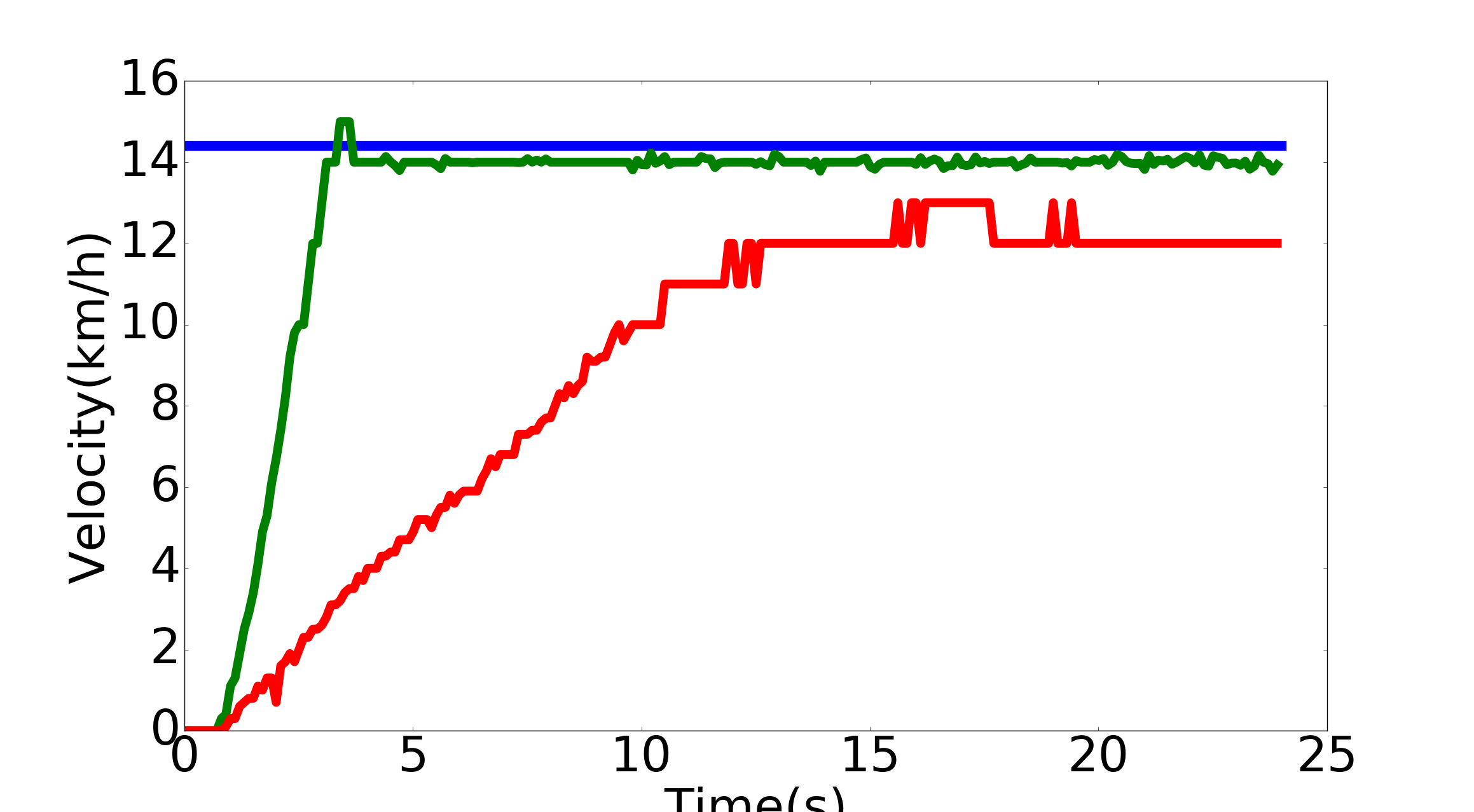}
        \caption{Performance over Varying\\Gradient-surface}
        \label{sp_gradient}
    \end{subfigure}
     \caption{Tracking set point velocity over different\\terrains}
     \label{set_point}
\end{figure}

We provide a reference velocity profile as a step input to the lower level control. The velocity to be tracked is $4m/s~ (14.4 km/hr)$ and the time-step given by the planning framework to achieve this velocity is $2 seconds$. Fig.\ref{sp_flat} shows the response of the vehicle to the step input on a flat surface and Fig.\ref{sp_gradient} shows the same on a varying gradient surface. Table\ref{tab:risetime} shows the rise time taken by the two control frameworks and it is seen that due to the shrinking domain control, the proposed controller reaches the velocity of $4m/s$ almost within the specified time of $2 seconds$. Whereas the PID controller not considering response time as a parameter, takes comparatively more time to reach the set-point.  

\subsubsection{Tracking a Continuously rising Velocity Profile provided by a planning framework}
\begin{figure}[H] 
    \begin{subfigure}[b]{0.48\linewidth}
        \includegraphics[width=\textwidth,height = 3.5cm]{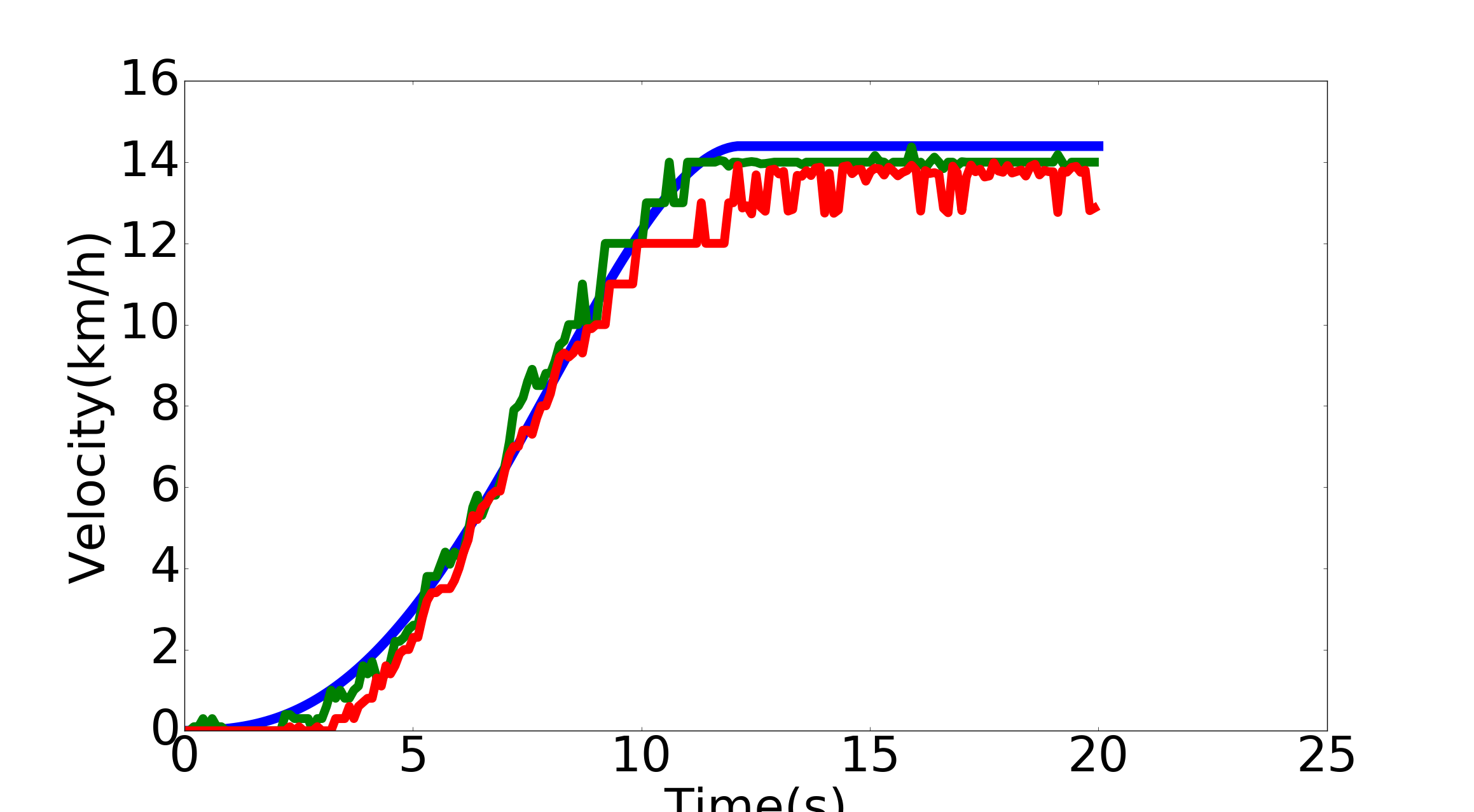}
        \caption{Performance over\\Flat-surface}
        \label{cr_flat}
    \end{subfigure}
    \begin{subfigure}[b]{0.48\linewidth}
        \includegraphics[width=\textwidth,height = 3.5cm]{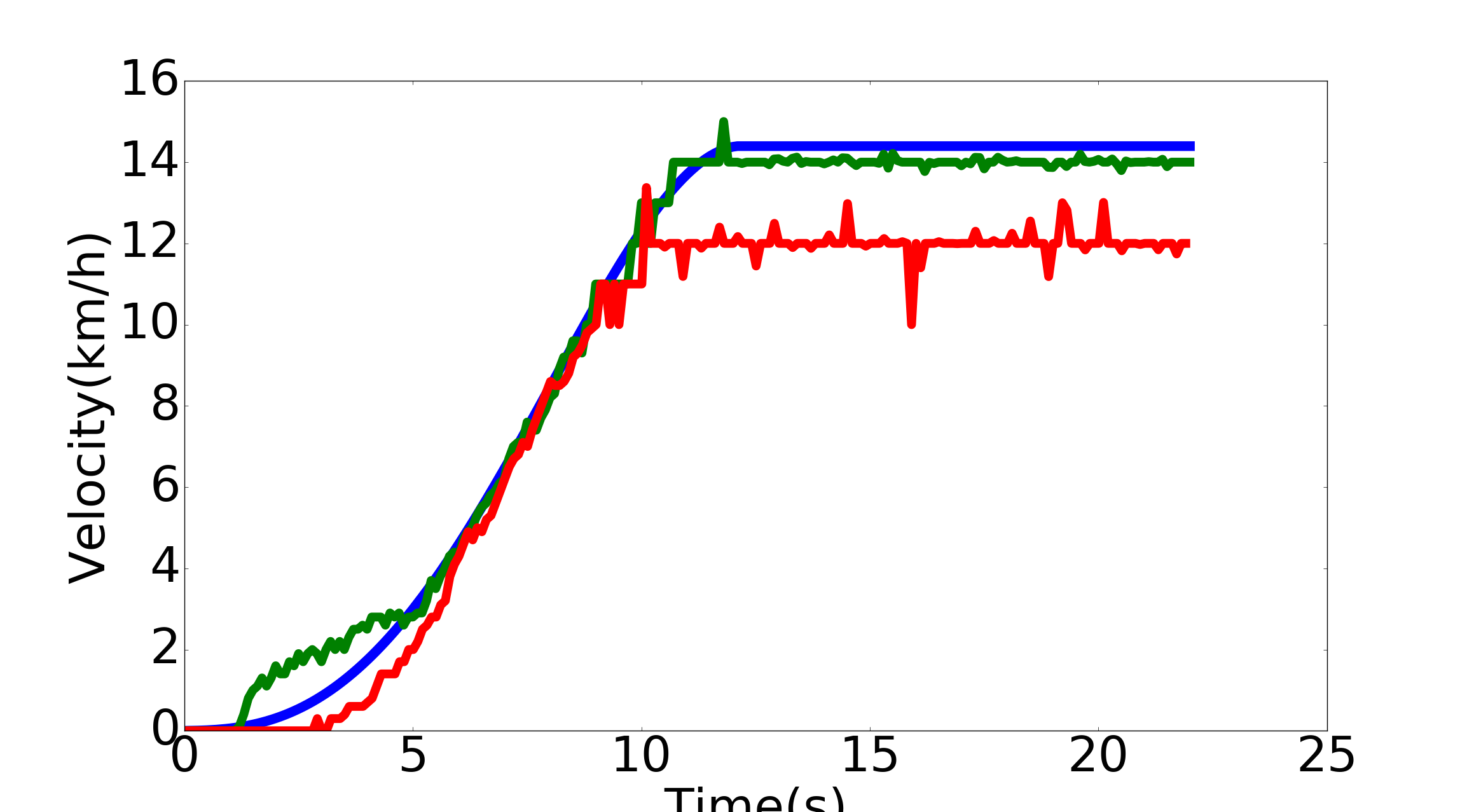}
        \caption{Performance over\\Varying Gradient-surface}
        \label{cr_gradient}
    \end{subfigure}
     \caption{Velocity Tracking of a gradually rising velocity profile over different terrains}
     \label{continuous_rise}
\end{figure}

Here, a continuous and a gradually rising velocity profile is given to the lower level control. The reference velocity profile increases to set point of $4m/s~(14.4 km/hr)$ to become constant. As seen in the Fig.\ref{cr_flat}, on a flat surface, both the PID controller and the proposed controller perform reasonably well. When tested on a varying gradient surface, however, we observe that the PID retains a steady state error, whereas the proposed controller has a comparatively lower error in velocity tracking.  This is visible in  Fig\ref{cr_gradient}.

\subsubsection{Tracking a Continuous Velocity profile from a planning framework in Stop-and-Go Scenarios}
\begin{figure}[H] 
    \begin{subfigure}[b]{0.48\linewidth}
        \includegraphics[width=\textwidth,height = 3.5cm]{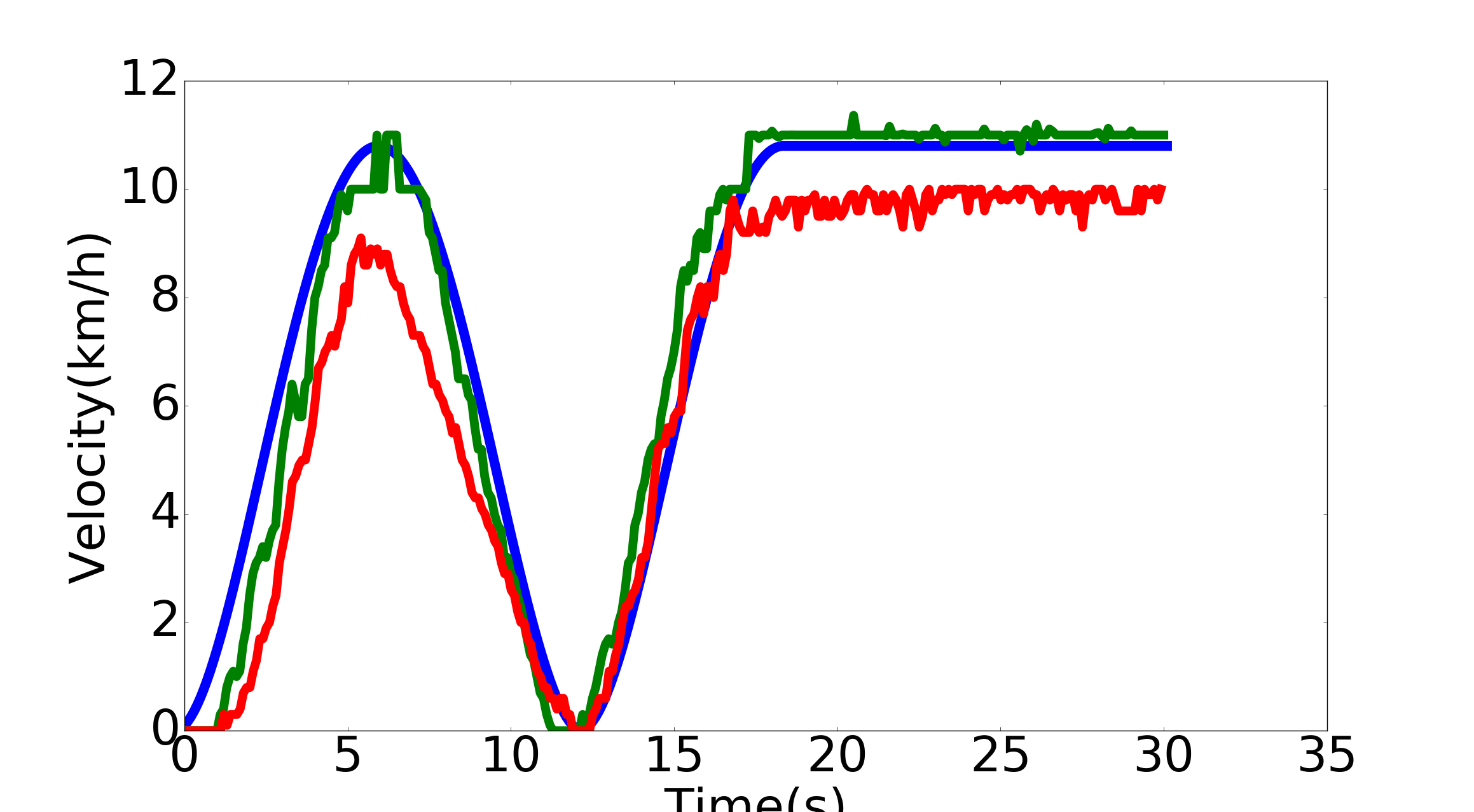}
        \caption{Performance over\\Flat-surface}
        \label{sng_flat}
    \end{subfigure}
    \begin{subfigure}[b]{0.48\linewidth}
        \includegraphics[width=\textwidth,height = 3.5cm]{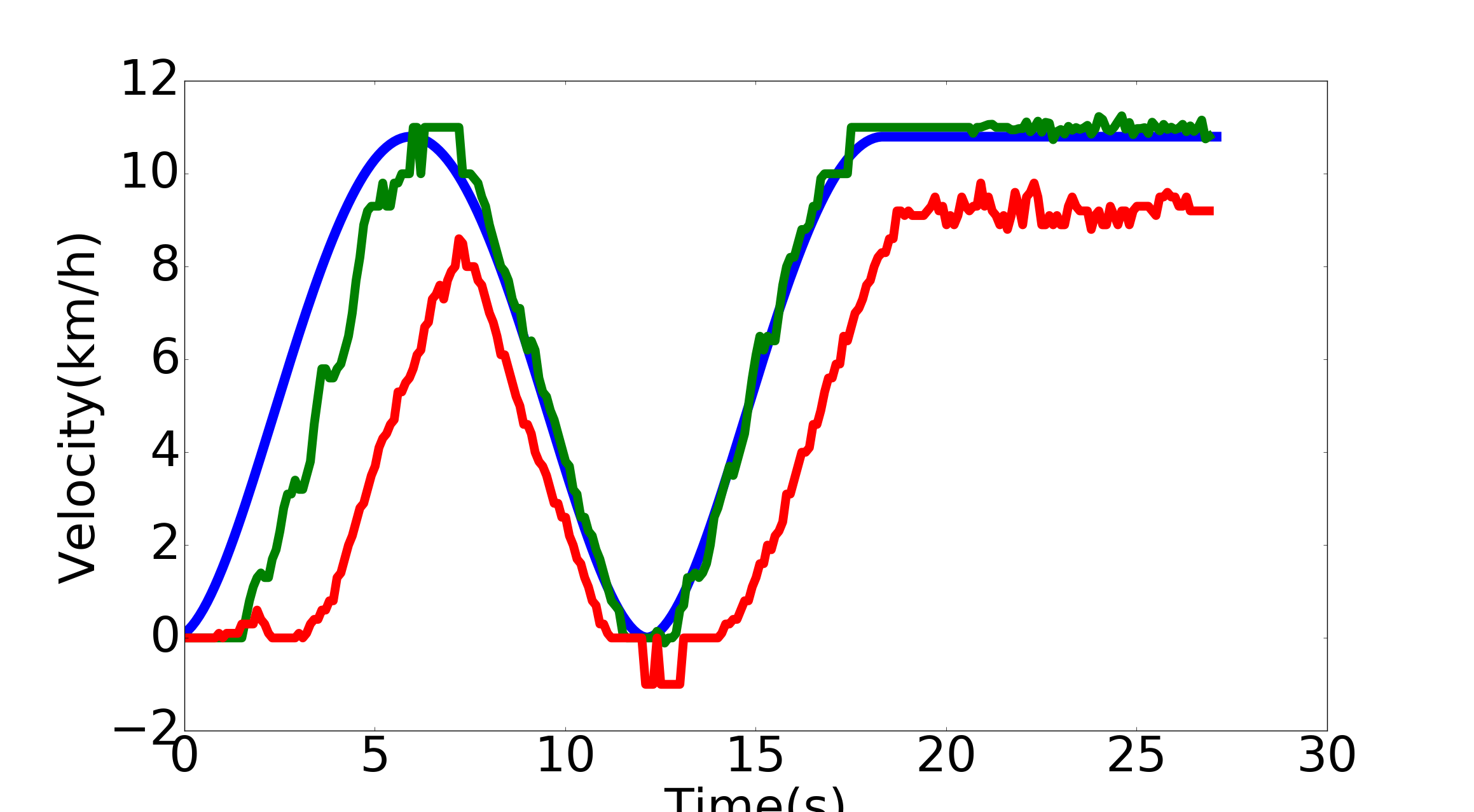}
        \caption{Performance over\\Varying Gradient-surface}
        \label{sng_gradient}
    \end{subfigure}
     \caption{Velocity tracking in a stop-and-go scenario over different terrains}
     \label{stopngo}
\end{figure}

Stop-and-go scenarios constantly occur in urban driving conditions and pose as dynamically challenging scenarios for an autonomous vehicle as both acceleration and braking come into picture. Due to the small inter-vehicle gap (during slow moving traffic), it is crucial for the vehicle's lower level control to follow the plan provided by the planning framework accurately within the given time constraints. Failing to do so may even result in an accident. In the results it is visible that the reference velocity profile rises to a velocity of $3m/s~(10.8 km/hr)$ and then falls to $0m/s$ and rises back to $3m/s$. As shown in Fig.\ref{sng_flat}, for the flat surface both the controllers track the profile reasonably well while barring the steady state error in case of PID controller. However, for varying gradient surfaces, the PID lags behind the reference profile as depicted in Fig.\ref{sng_gradient} which increases the steady state error. This lag also increases the RMS error which is tabulated in Table\ref{tab:rms error}.     

The proposed control strategy was incorporated in our autonomous electric vehicle. The scene was made in a similar way to emulate the urban driving scenario and the vehicle was tested in the conditions where a stop and go strategy has to be implemented as shown in the video below. The vehicle uses the LiDAR information to generate the occupancy map of the environment. This occupancy map is used by the high level TSCC\cite{2017arXiv171204965B} planner to generate collision free control velocities . As shown in the video our vehicle was successfully able to achieve and maintain the velocity in the gradient as well as stop at the situation when required and was able to follow the vehicle in the front.
The detailed demonstration can be found at (\href{https://youtu.be/Yf4F0dvkwQE}{https://youtu.be/Yf4F0dvkwQE})

\subsection{Empirical Analysis}

\subsubsection{Rise Time}
In order to assess and compare the reactivity of the proposed control framework, the rise time was calculated for the given step input. Rise time is being considered as the time taken for the velocity value to reach from 10\% to 90\% of its final steady state value.  As seen in the Fig.\ref{set_point}, the proposed control framework proves to be highly reactive and gives a very short rise time which is always nearly equal to the time-step provided by the planning framework inputs. The conventional PID however provides a high rise time which is not feasible for reactive planning.

\begin{table}[H] 
  \begin{center}
    \caption{Rise Time comparison between control frameworks (in seconds)} \label{tab:risetime}
    \begin{tabular}{c|c|c|c}
      \toprule
      \textbf{Velocity Profile} & \textbf{Terrain} & \textbf{\thead{PID \\Controller}} & \textbf{\thead{Proposed \\Controller}} \\
      \midrule
      
      \multirow{2}{*}{Set-point} 
      
      & flat & 10.07 & 1.81 \\ 
      \cmidrule{2-4}
      
      & gradient & 13.60 & 2.05\\ 
      \bottomrule

	\end{tabular}
  \end{center}
\end{table}

\subsubsection{Root Mean Squared Error}

Empirical calculations are performed to derive the root mean square deviation of the response from the planning framework commands. The RMS deviation is derived from various dynamic conditions the vehicle was put through, such as transient velocity tracking over gradient and flat surface in continuous and stop-and-go scenario. The RMS deviation clearly shows that the proposed control framework provides a very robust performance which is maintained throughout the dynamic conditions irrespective of gradient and driving conditions in different scenarios.

\begin{table}[H] 
  \begin{center}
    \caption{Comparison of Root Mean Squared Error for control frameworks(in $km/hr$)} \label{tab:rms error}
    \begin{tabular}{c|c|c|c}
      \toprule
      \textbf{Velocity Profile} & \textbf{Terrain} & \textbf{\thead{PID \\Controller}} & \textbf{\thead{Proposed \\Controller}} \\
      \midrule
      
      \multirow{2}{*}{\thead{Continuously \\ Rising}}       
      & flat & 0.0635 & 0.0295 \\ 
      \cmidrule{2-4}      
      & gradient & 0.1201 & 0.0390\\
\midrule            
      \multirow{2}{*}{\thead{Continuous \\ Stop-and-Go}}       
      & flat & 0.0904 & 0.0467 \\ 
      \cmidrule{2-4}      
      & gradient & 0.2050 & 0.0708\\
      
      \bottomrule

	\end{tabular}
  \end{center}
\end{table}

\subsubsection{Steady State Error}
Empirical calculations are also performed to derive the steady state error for the different dynamic driving conditions which the vehicle was put through. The steady state error also shows that the proposed framework maintains its integrity with varying dynamic conditions and minimum steady state error is maintained throughout as compared to the PID controller.

\begin{table}[H] 
  \begin{center}
    \caption{ Comparison of Steady State Error for control framework (in $km/hr$)} \label{tab:ss error}
    \begin{tabular}{c|c|c|c}
      \toprule
      \textbf{Velocity Profile} & \textbf{Terrain} & \textbf{\thead{PID \\Controller}} & \textbf{\thead{Proposed \\Controller}} \\
      \midrule

      \multirow{2}{*}{Set-point}       
      & flat & 0.5091 & 0.3853 \\ 
      \cmidrule{2-4}      
      & gradient & 2.2026 & 0.3935\\
\midrule  
      \multirow{2}{*}{\thead{Continuously \\ Rising}}       
      & flat & 0.8994 & 0.3931 \\ 
      \cmidrule{2-4}      
      & gradient & 2.3945 & 0.3991\\
\midrule            
      \multirow{2}{*}{\thead{Continuous \\ Stop-and-Go}}       
      & flat & 1.0199 & 0.2127 \\ 
      \cmidrule{2-4}      
      & gradient & 1.7345 & 0.2024\\
      
      \bottomrule

	\end{tabular}
  \end{center}
\end{table}

The main advantages as shown in the results are achieved in the gradient control as well as the reaction time of the vehicle system. Capturing the non-linearity of the system through the vehicle dynamics and motor characteristics, it has been possible to achieve a response which is highly beneficial for the planning framework.

\section{Conclusion and Future Work} \label{section5}

It is evident that the proposed framework is able to provide reasonably better dynamic response as compared to conventional control methodologies. The Autonomous Vehicle is able to track the velocity profile over varying terrain gradients in real time. The approach focuses on the decrement of the computation efforts of the pipeline while not losing out on the robustness and performance measures. Additionally, the framework makes sure that the planning framework commands are executed in real time within the prescribed time steps leading to minimum lag in the pipeline from the control framework's end. 

Further work can be done in the same framework by including lateral vehicle dynamics in real time with time optimality. This can be performed by collecting more data for various dynamic conditions and deriving system responses. Understanding system response can lead to system identification and better integrated control frameworks which are coherent with the entire Autonomous Vehicle Pipeline. 
Including Lateral Vehicle dynamics in the systems can provide immense support in performing more complicated maneuvers for autonomous vehicles which are governed by lateral dynamics such as dynamic obstacle avoidance and lane merging at high speeds.

\bibliography{cdc}

\begin{thebibliography}{10}
\providecommand{\url}[1]{#1}
\csname url@rmstyle\endcsname
\providecommand{\newblock}{\relax}
\providecommand{\bibinfo}[2]{#2}
\providecommand\BIBentrySTDinterwordspacing{\spaceskip=0pt\relax}
\providecommand\BIBentryALTinterwordstretchfactor{4}
\providecommand\BIBentryALTinterwordspacing{\spaceskip=\fontdimen2\font plus
\BIBentryALTinterwordstretchfactor\fontdimen3\font minus
  \fontdimen4\font\relax}
\providecommand\BIBforeignlanguage[2]{{%
\expandafter\ifx\csname l@#1\endcsname\relax
\typeout{** WARNING: IEEEtran.bst: No hyphenation pattern has been}%
\typeout{** loaded for the language `#1'. Using the pattern for}%
\typeout{** the default language instead.}%
\else
\language=\csname l@#1\endcsname
\fi
#2}}

\bibitem{kuiper2007pac2002}
E.~Kuiper and J.~Van~Oosten, ``The pac2002 advanced handling tire model,''
  \emph{Vehicle system dynamics}, vol.~45, no.~S1, pp. 153--167, 2007.

\bibitem{801226}
M.~Persson, F.~Botling, E.~Hesslow, and R.~Johansson, ``Stop and go controller
  for adaptive cruise control,'' in \emph{Proceedings of the 1999 IEEE
  International Conference on Control Applications (Cat. No.99CH36328)},
  vol.~2, 1999, pp. 1692--1697 vol. 2.

\bibitem{4105943}
J.~J. Martinez and C.~C. de~Wit, ``A safe longitudinal control for adaptive
  cruise control and stop-and-go scenarios,'' \emph{IEEE Transactions on
  Control Systems Technology}, vol.~15, no.~2, pp. 246--258, March 2007.

\bibitem{choi2002hybrid}
H.-C. Choi and S.-K. Hong, ``Hybrid control for longitudinal speed and traction
  of vehicles,'' in \emph{IECON 02 [Industrial Electronics Society, IEEE 2002
  28th Annual Conference of the]}, vol.~2.\hskip 1em plus 0.5em minus
  0.4em\relax IEEE, 2002, pp. 1675--1680.

\bibitem{antsaklis2000brief}
P.~J. Antsaklis, ``A brief introduction to the theory and applications of
  hybrid systems,'' in \emph{Proc IEEE, Special Issue on Hybrid Systems: Theory
  and Applications}.\hskip 1em plus 0.5em minus 0.4em\relax Citeseer, 2000.

\bibitem{decarlo2000perspectives}
R.~A. DeCarlo, M.~S. Branicky, S.~Pettersson, and B.~Lennartson, ``Perspectives
  and results on the stability and stabilizability of hybrid systems,''
  \emph{Proceedings of the IEEE}, vol.~88, no.~7, pp. 1069--1082, 2000.

\bibitem{1227578}
M.~A. Goodrich and E.~R. Boer, ``Model-based human-centered task automation: a
  case study in acc system design,'' \emph{IEEE Transactions on Systems, Man,
  and Cybernetics - Part A: Systems and Humans}, vol.~33, no.~3, pp. 325--336,
  May 2003.

\bibitem{shakouri2011adaptive}
P.~Shakouri, A.~Ordys, D.~S. Laila, and M.~Askari, ``Adaptive cruise control
  system: comparing gain-scheduling pi and lq controllers,'' \emph{IFAC
  Proceedings Volumes}, vol.~44, no.~1, pp. 12\,964--12\,969, 2011.

\bibitem{7795572}
F.~A. Mullakkal-Babu, M.~Wang, B.~van Arem, and R.~Happee, ``Design and
  analysis of full range adaptive cruise control with integrated collision a
  voidance strategy,'' in \emph{2016 IEEE 19th International Conference on
  Intelligent Transportation Systems (ITSC)}, Nov 2016, pp. 308--315.

\bibitem{5625228}
J.~Villagrá, V.~Milanés, J.~Perez, and C.~González, ``Model-free control
  techniques for stop amp; go systems,'' in \emph{13th International IEEE
  Conference on Intelligent Transportation Systems}, Sept 2010, pp. 1899--1904.

\bibitem{7531726}
X.~Lu and J.~Fei, ``Velocity tracking control of wheeled mobile robots by fuzzy
  adaptive iterative learning control,'' in \emph{2016 Chinese Control and
  Decision Conference (CCDC)}, May 2016, pp. 4242--4247.

\bibitem{kim2016time}
H.~Kim, D.~Kim, I.~Shu, and K.~Yi, ``Time-varying parameter adaptive vehicle
  speed control,'' \emph{IEEE Transactions on Vehicular Technology}, vol.~65,
  no.~2, pp. 581--588, 2016.

\bibitem{chen2017adaptive}
H.~Chen, L.~Liu, Y.~Zhang, and Y.~Tian, ``Adaptive speed control of autonomous
  vehicle under changing operation conditions,'' in \emph{Control Conference
  (CCC), 2017 36th Chinese}.\hskip 1em plus 0.5em minus 0.4em\relax IEEE, 2017,
  pp. 3522--3526.

\bibitem{chen2011adaptive}
Y.~Chen and J.~Wang, ``Adaptive vehicle speed control with input injections for
  longitudinal motion independent road frictional condition estimation,''
  \emph{IEEE Transactions on Vehicular Technology}, vol.~60, no.~3, pp.
  839--848, 2011.

\bibitem{moon2009design}
S.~Moon, I.~Moon, and K.~Yi, ``Design, tuning, and evaluation of a full-range
  adaptive cruise control system with collision avoidance,'' \emph{Control
  Engineering Practice}, vol.~17, no.~4, pp. 442--455, 2009.

\bibitem{rajamani2011vehicle}
R.~Rajamani, \emph{Vehicle dynamics and control}.\hskip 1em plus 0.5em minus
  0.4em\relax Springer Science \& Business Media, 2011.

\bibitem{gillespie1992fundamentals}
T.~D. Gillespie, ``Fundamentals of vehicle dynamics. warrendale, pa: Society of
  automotive engineers,'' 1992.

\bibitem{mohan1995power}
N.~Mohan and T.~M. Mohan, \emph{Power electronics}.\hskip 1em plus 0.5em minus
  0.4em\relax John wiley \& sons New York, 1995, vol.~3.

\bibitem{dubey2002fundamentals}
G.~K. Dubey, \emph{Fundamentals of electrical drives}.\hskip 1em plus 0.5em
  minus 0.4em\relax CRC Press, 2002.

\bibitem{fitzgerald2003electric}
A.~E. Fitzgerald, C.~Kingsley, S.~D. Umans, and B.~James, \emph{Electric
  machinery}.\hskip 1em plus 0.5em minus 0.4em\relax McGraw-Hill New York,
  2003, vol.~5.

\bibitem{del2010automotive}
L.~Del~Re, F.~Allg{\"o}wer, L.~Glielmo, C.~Guardiola, and I.~Kolmanovsky,
  \emph{Automotive model predictive control: models, methods and
  applications}.\hskip 1em plus 0.5em minus 0.4em\relax Springer, 2010, vol.
  402.

\bibitem{2017arXiv171204965B}
M.~{Babu}, Y.~{Oza}, A.~K. {Singh}, K.~{Madhava Krishna}, and S.~{Medasani},
  ``{Model Predictive Control for Autonomous Driving Based on Time Scaled
  Collision Cone},'' \emph{ArXiv e-prints}, Dec. 2017.

\end{thebibliography}
\bibliographystyle{IEEEtran}

\end{document}